\documentclass{article}
\usepackage{spconf,amsmath,graphicx}
\usepackage{amsfonts}
\usepackage{enumitem}
\usepackage{url}
\setlist{nosep, leftmargin=14pt}

\usepackage{mwe} % to get dummy images
\usepackage{hyperref}
\usepackage{cite}

\providecommand{\norm}[1]{\ensuremath{\left\lVert #1 \right\rVert}}

\title{SLAM-AGS: Slide-Label Aware Multi-Task Pretraining Using Adaptive Gradient Surgery in Computational Cytology}
\name{Marco Acerbis, Swarnadip Chatterjee, Christophe Avenel, Joakim Lindblad}
\address{Center for Image Analysis, Department of Information Technology, Uppsala University, Sweden}
\begin{document}
%\ninept
%
\maketitle
\begin{abstract}
Computational cytology faces two major challenges: i) instance-level labels are unreliable and prohibitively costly to obtain, ii) witness rates are extremely low. We propose \textbf{SLAM-AGS}, a \textbf{S}lide-\textbf{L}abel–\textbf{A}ware \textbf{M}ultitask pretraining framework that jointly optimizes (i) a weakly supervised similarity objective on slide-negative patches and (ii) a self-supervised contrastive objective on slide-positive patches, yielding stronger performance on downstream tasks. To stabilize learning, we apply \textbf{A}daptive \textbf{G}radient \textbf{S}urgery to tackle conflicting task gradients and prevent model collapse. We integrate the pretrained encoder into an attention-based Multiple Instance Learning aggregator for bag-level prediction and attention-guided retrieval of the most abnormal instances in a bag. On a publicly available bone-marrow cytology dataset, with simulated witness rates from 10\% down to 0.5\%, SLAM-AGS improves bag-level F1-Score and Top 400 positive cell retrieval over other pretraining methods, with the largest gains at low witness rates, showing that resolving gradient interference enables stable pretraining and better performance on downstream tasks. To facilitate reproducibility, we share our complete implementation and evaluation framework as open source: \href{https://github.com/Ace95/SLAM-AGS}{https://github.com/Ace95/SLAM-AGS}.
\end{abstract}


\section{Introduction}
Deep learning has demonstrated state-of-the-art performance in medical image analysis, particularly for cancer detection on histopathology whole-slide images (WSIs). For gigapixel WSIs, obtaining exhaustive patch-level labels is prohibitively costly, making fully supervised training of patch-wise predictors infeasible. A common solution is to adopt weak supervision, i.e. to use slide-level labels only. In addition, the witness rates (positive patches per positive slide) are often extremely low (less than 1\%) in cytology WSIs. At such low witness rates, supervision is dominated by negative instances, leading to overfitting and poor generalization to rare positives \cite{its2clr}.

%In weakly labeled datasets, where only slide-level annotations are available, the number of positive instances per sample (witness rate) is often low. This hampers learning, as networks tend to overfit the dominant negative class and fail to generalize to rare positive instances.

\begin{figure}[htb!]
\begin{minipage}[b]{1.0\linewidth}
  \centering
  \centerline{\includegraphics[width=\linewidth]{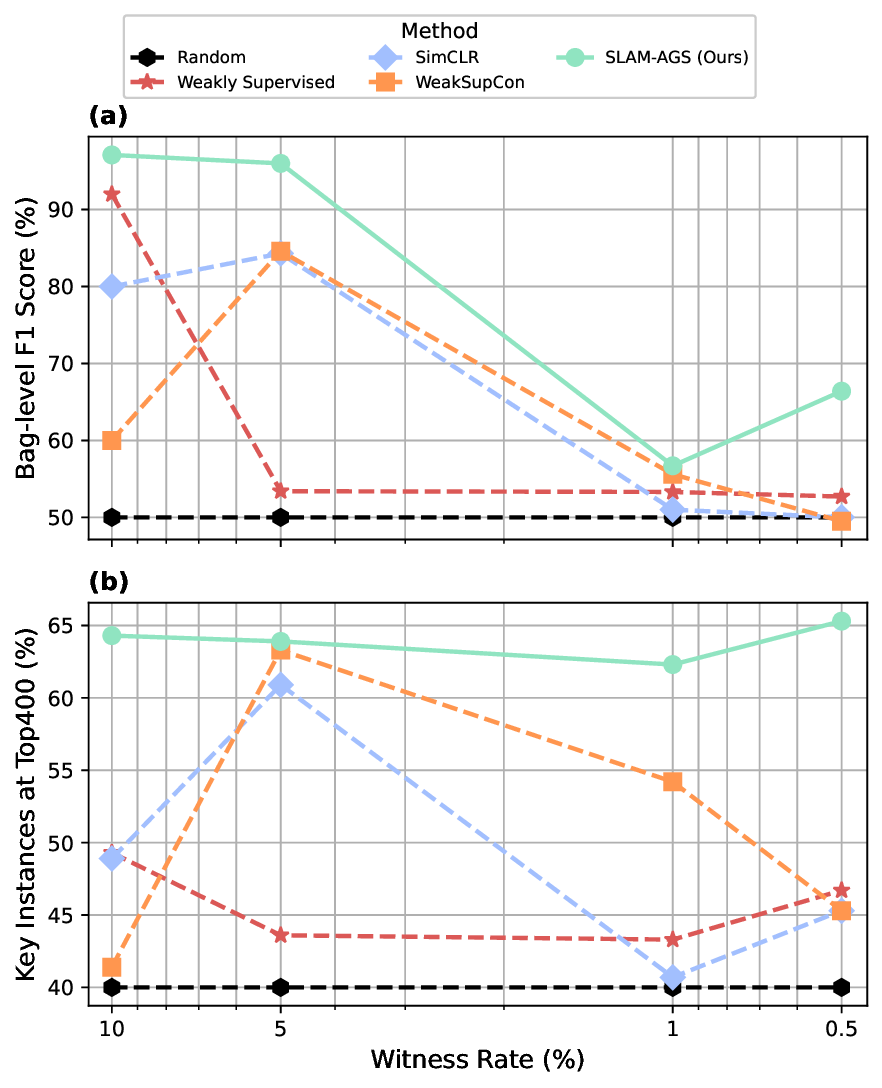}}
\end{minipage}
\caption{(a) F1-scores for the bag classification task across decreasing witness rates. (b) Recall at Top 400 patches (corresponding to top 40\% of instances) of positive bags.}
\label{fig:res}
\end{figure}

With only slide-level labels available, Multiple Instance Learning (MIL) has been widely adopted for histopathological analysis\cite{abmil, DTFD, PAMIL}. MIL frameworks treat each WSI as a bag of instances (patches) and learn to infer the bag label from instance-level features using an aggregator, without requiring explicit patch-level annotations. Attention-based MIL models offer interpretability by further identifying the most relevant patches contributing to a positive diagnosis, via attention scores. However, MIL approaches still face difficulties in scenarios with very low witness rates or limited training data, as the instance-level discrimination remains weakly supervised.
Self-supervised representation learning has emerged as a promising complementary method, enabling models to learn robust visual features from unlabeled data. Yet, standard self-supervised pretraining frameworks do not leverage available weak labels, leaving the potential of bag-level supervision untapped.
Recent work \cite{weaksupcon} introduced a multitask method that combines self-supervised and weakly supervised objectives, yielding promising results. However, combining heterogeneous objectives in a multi-task setting often leads to gradient conflicts, which can destabilize training and limit the benefit of weak supervision.\\[-0.3ex]

Shifting our focus to cytology, the adoption of deep learning methods has progressed more slowly compared to histopathology. Cytology WSI analysis introduces additional challenges, including higher resolution requirements, limited availability of public datasets, large intra-class variability, and the difficulty of obtaining reliable cell-level annotations. Furthermore, abnormal cells often appear sparsely across samples, exacerbating the low witness rate problem. In computational cytology for malignancy detection, two main objectives are crucial: (i) performing bag-level classification to identify affected patients and (ii) detecting key patches within each bag corresponding to abnormal cells (instance-level detection).\\[-0.3ex]

%Compared to histopathology, cytopathology has experienced slower progress in adopting deep learning methods. Cytological analysis poses additional challenges, including higher resolution requirement, limited publicly available datasets, large intra-class variability, and difficulties in obtaining reliable cell-level annotations. Furthermore, abnormal cells may appear sparsely across samples, exacerbating the low witness rate problem. In digital cytology for cancer detection, two main objectives are crucial: (i) identifying patients developing cancer (bag-level classification) and (ii) detecting key instances within each bag corresponding to abnormal cells (instance-level detection). \\

To address the issue of poor performance of weakly supervised methods when facing very low witness rates, we propose a Slide-Label Aware Multitask pretraining approach that builds on WeakSupCon \cite{weaksupcon} and incorporates Adaptive Gradient Surgery to improve training stability.
In this work we:
\begin{itemize}
    \item Adapt and evaluate existing Multi-Task pretraining techniques in cytology settings. In particular, we consider small datasets and low witness rates;
    \item Implement Adaptive Gradient Surgery during training to manage conflicts between gradients;
    \item Introduce a renewed MIL pipeline for both bag- and instance-level classification in computational cytology. 
\end{itemize}

\section{Related work}

\subsection{Multiple Instance Learning}
In WSI analysis, fully supervised patch-level training is often impractical due to gigapixel scale and the cost of exhaustive annotation.  Multiple Instance Learning (MIL) addresses this by treating each WSI as a bag of patch instances with a slide-level label and training an aggregator to infer the bag label from instance features. Since the goal most often is not only to classify the bags (e.g., detect the presence of malignancy in a WSI) but also to identify the key patches (e.g., malignant regions or cells), several methods have been proposed to highlight positive instances within each bag. Notable examples include Attention-Based MIL (ABMIL)\cite{abmil} and DTFD-MIL\cite{DTFD}. Prototype Attention-Based MIL (PAMIL)\cite{PAMIL} further improves interpretability by learning class-specific prototypes that assign class-specific attention scores to patches, indicating their contribution to the overall bag prediction.

For scenarios with extremely sparse key instances, self-supervised one-class representations can effectively retrieve abnormal cells when MIL models struggle\cite{swarnadip}. However, relying only on normal slides and lacking an aggregator, this approach falls behind other methods at higher witness rates.

%For scenarios with extremely sparse key instances, self-supervised one-class representations have been shown to be able to effectively retrieve abnormal cells when MIL models struggle\cite{swarnadip}.
%under extremely low witness rates.
%However, relying on normal slides only, this approach falls behind other methods for higher witness rates.

\subsection{Pretraining}
A major challenge for MIL approaches is to effectively pretrain the encoder. Supervised methods tend to fail when the witness rate decreases, as positive instances become too sparse to guide effective learning.

Self-supervised contrastive pretraining (e.g., SimCLR \cite{simclr}, MoCo v3 \cite{mocov3}) provides an alternative by learning label-free, model-agnostic representations with strong cross-domain generalization. However, they do not exploit weak labels, which are commonly available in MIL settings. Recent supervised contrastive objectives, like SupCon \cite{supcon}, bring labels into contrastive learning and, when used for pretraining or fine-tuning, often outperform Cross-Entropy-based supervised training and purely self-supervised contrastive methods, highlighting the potential benefits of combining the two paradigms.
Motivated by this, weak-label-aware variants adapt contrastive pretraining to MIL. WeakSupCon \cite{weaksupcon} integrates weak supervision with contrastive representation learning, forming a multi-task pretraining scheme. In particular, slide-negative patches that share the same label participate in a supervised-like task, whereas slide-positive patches, which may contain both normal and abnormal cells, are used to optimize a self-supervised contrastive objective.

\subsection{Multi-Task Learning Optimization}
Multi-Task Learning improves generalization by jointly optimizing related objectives. Yet training a shared encoder with multiple losses can induce gradient conflicts and task-scale imbalance, destabilizing optimization. To address this, MGDA \cite{mgda} constrains updates to directions that do not harm any task, while PCGrad \cite{pcgrad} subtracts conflicting gradient projections to smooth convergence—though this may overly distort directions and magnitudes, weakening the learning signal. Recent methods \cite{CAGrad, IMTL} adapt gradient corrections on-the-fly to (i) reduce conflicts and (ii) preserve informative update scales, yielding more stable training.

\section{Method}

We propose SLAM-AGS: a multi-task pretraining framework, by adapting WeakSupCon into a slide-label aware multitask pretraining setup and using adaptive gradient surgery for label-efficient representation learning. 

Slide-negative patches extracted from healthy patients are treated in a supervised manner, as they share the same known (negative) instance label, encouraging compact representation in the feature space. Following WeakSupCon, we implement a Similarity Loss derived from SupCon Loss for slide-negative patches:
\vspace*{-2mm}
\begin{equation}
\mathcal{L}_{\texttt{Similarity}}
=  \frac{-1}{\lvert \texttt{\textbf{Neg}} \rvert}
\sum_{\substack{i,j \in \texttt{\textbf{Neg}}\\ i \neq j}}
\frac{z_i \cdot z_j}{\tau} \, ,
\vspace*{-1mm}
\end{equation}
where $\texttt{\textbf{Neg}}$ represent the set of all slide-negative patches in a batch and $\tau$ is temperature scaling hyperparameter.

Slide-positive patches, which may contain both normal and abnormal cells, are used to optimize the SimCLR\cite{simclr} objective:
%%%%%%%%%%%%%%%%%%%%%%%%%%%%%%%%%
\begin{equation}
\mathcal{L}_{\texttt{SimCLR}}
= \frac{-1}{2\lvert\texttt{\textbf{Pos}}\rvert}
\sum_{(i,j)\in\mathcal{P}_{\texttt{\textbf{Pos}}}}\!\!\!\!\!\!\log
\frac{
    \exp(\texttt{sim}(z_i,z_j)/\tau)
}{
    \sum\limits_{k\in\mathcal{V}_{\texttt{\textbf{Pos}}}\setminus\{i\}}
    \exp(\texttt{sim}(z_i,z_k)/\tau)
} .
\end{equation}
where $\texttt{\textbf{Pos}}$ represents the set of slide-positive patches in the batch;
$\mathcal{V}_{\texttt{\textbf{Pos}}}$ is the set of $2\lvert\texttt{\textbf{Pos}}\rvert$ views (two stochastic augmentations per patch);
$\mathcal{P}_{\texttt{\textbf{Pos}}}\!\!
:=\!\!\{(i,j)\!\!\mid \!\!i,j \!\in \mathcal{V}_{\texttt{\textbf{Pos}}},\, \!\!i \!\!\neq \!\!j,\,
i \text{ and } j \text{ are views of the same patch}\}$ is the set of ordered positive pairs;
\texttt{sim}$(\cdot,\cdot)$ is the cosine similarity;
and $z_i$ denotes the (projected) embedding of view $i$.
% \begin{equation}
% \mathcal{L}_{\texttt{SimCLR}}
% = \frac{-1}{\left|\mathcal {\texttt{\textbf{Pos}}}\right|}
% \sum_{i \in \mathcal{\texttt{\textbf{Pos}}}}
% \log \frac{\exp\!\left(\mathrm{\texttt{sim}}(z_i, z_{p(i)})/\tau\right)}
% {\sum_{k=1}^{2N} I_{k \neq i} \exp(sim(z_i,z_k)/\tau)} \, .
% \end{equation}
% where $\texttt{\textbf{Pos}}$ represents the set of slide-positive patches in a batch, $sim(\cdot,\cdot)$ is the pairwise cosine similarity, $N$ is the batch size, $z_i$ and $z_{p(i)}$ are the embeddings of the augmented views of patch $i$; and $\{z_k\}$ are the embeddings of all other slide-positive patches in the batch.
%%%%%%%%%%%%%%%%%%%%%%%%%%%%%%%%%%
% \begin{equation}
% \mathcal{L}_{\texttt{SimCLR}} = -\!\!\!\!\sum_{i \in \texttt{\textbf{Pos}}}\!\!\log \frac{\exp(sim(z_i,z_{p(i)})/\tau)}{\sum_{k=1}^{2N} I_{k \neq i} \exp(sim(z_i,z_k)/\tau)},
% \end{equation}
% where $\texttt{\textbf{Pos}}$ represents the set of slide-positive patches in a batch, $sim(\cdot,\cdot)$ is the pairwise cosine similarity, $N$ is the batch size, $z_i$ denotes the (projected) embedding of patch $i$; $z_{p(i)}$ is the embedding of an augmented view of the same patch (the positive); and $\{z_k\}$ are the embeddings of all other views from slide-positive patches in the batch.

The overall objective is defined as a combination of both losses:
\begin{equation}
\mathcal{L} = \mathcal{L}_{\texttt{Similarity}} + \mathcal{L}_{\texttt{SimCLR}}.
\end{equation}

Although our setup involves only two objectives, gradient conflicts and imbalanced magnitudes may destabilize optimization. To address this, we adopt the PCGrad algorithm, which projects each task gradient onto the normal plane of other task gradients when negative cosine similarity indicates conflict. This projection removes destructive interference between tasks. However, the projection can reduce the effective gradient magnitude, leading to slower convergence. 

\paragraph*{Gradient rescaling}
To mitigate the negative effects of the gradient projections, we introduce a gradient rescaling step. We first compute the total gradient norm before applying PCGrad:
\vspace*{-1mm}
\begin{equation}
    \norm{\mathbf{g}_{\text{sum}}} = \norm{\mathbf{g}_1 + \mathbf{g}_2},
    % \qquad
    % \norm{ \mathbf{g}_{\text{sum}} }.
\end{equation}
Then, we compute the total gradient norm after applying PCGrad:
\vspace*{-1mm}
\begin{equation}
    \norm{\mathbf{g}_{\text{pc}}} = \norm{\mathbf{g}_{\text{pc},1} + \mathbf{g}_{\text{pc},2}},
    % \qquad
    % \norm{ \mathbf{g}_{\text{pc}} }\,.
\end{equation}
If the projected gradient norm is smaller than the original one, we rescale it to preserve the overall magnitude:
\begin{equation}
    \text{if } 
    \norm{ \mathbf{g}_{\text{pc}} }
    < 
    \norm{ \mathbf{g}_{\text{sum}} },
    \quad
    \mathbf{g}_{\text{pc}} \leftarrow 
    \mathbf{g}_{\text{pc}} \cdot
    \frac{\norm{ \mathbf{g}_{\text{sum}} }}
         {\norm{ \mathbf{g}_{\text{pc}} }}\,.
\end{equation}
This preserves the overall update scale while maintaining non-conflicting directions. Empirically, as shown in the experimental results in Section \ref{sec:results}, this projection–rescaling step stabilizes optimization and harmonizes updates between the two objectives.

\section{Experiments}

\begin{figure}[tb]
\begin{minipage}[b]{.24\linewidth}
  \centering
  \includegraphics[width=2.0cm]{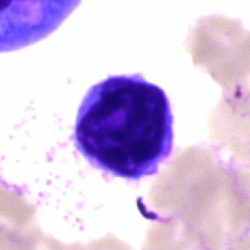}
  (a)
\end{minipage}
\hfill
\begin{minipage}[b]{0.24\linewidth}
  \centering
  \includegraphics[width=2.0cm]{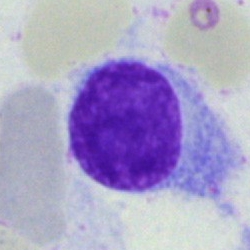}
  (b)
\end{minipage}
\begin{minipage}[b]{0.24\linewidth}
  \centering
  \includegraphics[width=2.0cm]{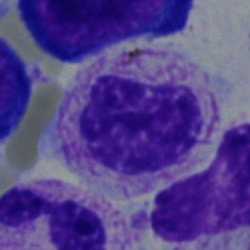}
  (c)
\end{minipage}
\begin{minipage}[b]{0.24\linewidth}
  \centering
  \includegraphics[width=2.0cm]{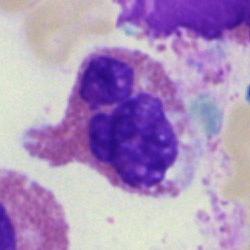}
  (d)
\end{minipage}
\caption{Examples of images extracted from the Bone Marrow Cytology dataset: (a) lymphocyte, (b) hairy cell, (c) promyelocyte, (d) basophil.}
\label{fig:pkgbm_examples}
\end{figure}
\subsection{Dataset}
Due to the lack of cytology datasets with reliable instance labels and controlled witness rates, we use cell images from the \textit{Bone Marrow Cytology} dataset\cite{pkgbm} (different hematologic cell types). Images come from \textit{May-Grünwald-Giemsa/Pappenheim}–stained WSIs (brightfield). To simulate an MIL setting with positive/negative bags, we label \textit{Lymphocytes} (the majority class) as \textit{``normal''} and merge seven minority classes into a single \textit{``abnormal''} class. We create abnormal bags by injecting a controlled fraction of abnormal instances into otherwise normal bags to match the target witness rates.  The distribution of normal and abnormal cells in the training and test sets is summarized in Table \ref{tab:pkgbm_dataset}, while Figure \ref{fig:pkgbm_examples} show some examples of images extracted from the dataset. Patches are grouped into bags of 1,000 images, with variable witness rates of 10\%, 5\%, 1\%, and 0.5\%.

\begin{table}[h]
\centering
\begin{tabular}{c|c|c}
\textbf{Cell Type} & \textbf{Train} & \textbf{Test} \\
\hline
Normal & 18,369 & 7,873 \\
Abnormal & 910 & 369 \\
\end{tabular}
\caption{Distribution of ``normal'' and ``abnormal'' cells in the Bone Marrow Cytology dataset.}
\label{tab:pkgbm_dataset}
\end{table}

\subsection{Experimental Design}
The experiments are designed to evaluate the impact of different pretraining strategies on MIL performance under varying witness rates. All methods pretrain a ResNet-18 encoder\cite{resnet18}. As a baseline, we use a Weakly Supervised method, in which patch instances inherit their slide label and a ResNet-18 encoder is trained on these propagated labels with a cross-entropy loss. For both train and test, positive bags are matched to the target witness rate. The pretrained encoders then provide embeddings to train the PAMIL aggregator. 

Model performance is assessed using two metrics: the F1-score, measuring bag-level classification performance, and Recall@400, measuring instance-level detection performance, i.e., the average number of correctly identified key instances among the top 400 patches, relative to the total number of key instances, where patches (instances) are ranked by their attention scores toward the positive class in positive bags. Each test is performed with five different random seeds and we report mean and standard deviation of the respective performance measures over these five runs.

\subsection{Implementation Details}
%Pretraining is conducted on the C3SE Alvis (NAISS) computational cluster equipped with NVIDIA A100 GPUs (40 GB memory). Testing is performed on a local workstation with 
%an Intel\textsuperscript{\textregistered} Core\textsuperscript{TM} i9-9940X CPU, running Gentoo Linux 6.6.67, Python 3.12.8, and PyTorch 2.5.1+cu124, using 
%an NVIDIA GeForce RTX 4090 GPU (24 GB memory).

Each model is trained for 150 epochs using a cosine annealing learning rate scheduler, with a warm-up phase of 10 epochs and an initial learning rate of 0.001. The batch size is set to 256, with each batch containing an equal number of positive and negative patches. Each training set consists of 18 bags, equally divided into positive and negative samples. Similarly, each test set contains a total of 6 bags, with 3 positive and 3 negative samples.

\begin{table}[h]
\caption{Bag-level performance (F1-score) for different witness rates (WR).}
\smallskip
\centering
\resizebox{\columnwidth}{!}{%
\begin{tabular}{c|cccc}
\hline
\textbf{Method} & \textbf{10\% WR} & \textbf{5\% WR} & \textbf{1\% WR} & \textbf{0.5\% WR} \\
\hline
Weakly Supervised & 92.0 $\pm$ 11.0 & 53.4 $\pm$ 29.8 & 53.3 $\pm$ 18.3 & 52.7 $\pm$ 19.2\\
SimCLR & 80.0 $\pm$ 11.2 & 84.3 $\pm$ 10.3 & 51.0 $\pm$ 17.5 & 50.0 $\pm$ 9.7 \\
WeakSupCon & 60.0 $\pm$ 14.9 & 84.6 $\pm$ 2.6 & 55.6 $\pm$ 9.6 & 49.5 $\pm$ 9.7 \\
SLAM-AGS w/o rescaling & 66.0 $\pm$ 10.6 & 60.0 $\pm$ 9.1 & 54.1 $\pm$ 16.9 & 46.0	$\pm$ 18.9 \\ 
SLAM-AGS (Ours) & \textbf{97.1} $\pm$ 6.4 & \textbf{96.0} $\pm$ 8.9 & \textbf{56.7} $\pm$ 14.9 & \textbf{66.4} $\pm$ 6.3 \\
\hline
\end{tabular}}
\label{tab:bag_f1}
\caption{Key instance detection performance (average number of correct key instances among Top 400) for different witness rates (WR).}
\centering
\resizebox{\columnwidth}{!}{%
\begin{tabular}{c|cccc}
\hline
\textbf{Method} & \textbf{10\% WR} & \textbf{5\% WR} & \textbf{1\% WR} & \textbf{0.5\% WR} \\
\hline
Weakly Supervised & 49.3 $\pm$ 15.7 & 43.6 $\pm$ 21.8 & 43.3 $\pm$ 3.3 & 46.7 $\pm$ 12.5 \\
SimCLR & 48.9 $\pm$ 11.4 & 60.9 $\pm$ 14.1 & 40.7 $\pm$ 22.4 & 45.3 $\pm$ 15.9 \\
WeakSupCon & 41.4 $\pm$ 3.0 & 63.3 $\pm$ 5.7 & 54.2 $\pm$ 10.0 & 45.3 $\pm$ 14.5 \\
SLAM-AGS w/o rescaling & 50.7 $\pm$ 13.9 & 43.7 $\pm$ 14.9 & 47.3 $\pm$ 13.2 & 46.7	$\pm$ 19.4\\
SLAM-AGS (Ours) & \textbf{64.3} $\pm$ 7.3 & \textbf{63.9} $\pm$ 8.8 & \textbf{62.0} $\pm$ 19.4 & \textbf{65.3} $\pm$ 11.9 \\
\hline
\end{tabular}}

\label{tab:top400}
\end{table}

\subsection{Results} \label{sec:results}
Table \ref{tab:bag_f1} and Figure~\ref{fig:res}(a) reports the F1-scores for the bag classification task. The proposed method achieves the highest performance across all witness rates. As expected, performance decreases for all methods as the witness rate diminishes, indicating that a weaker positive signal from fewer key instances makes it more difficult for the aggregator to distinguish between positive and negative bags. Another important observation is the collapsing of the baseline method for low witness rates to near random performance.

Table \ref{tab:top400} and Figure~\ref{fig:res}(b) presents the results for key instance detection. Again, the proposed SLAM-AGS method consistently outperforms the others with an average detection rate above 60\%, demonstrating improved ability to identify relevant instances even at low witness rates. In particular, we observe that the SimCLR and WeakSupCon pretrained models tend toward a random distribution of instances in positive bags (with a near random Top-400 detection rate of slightly over 40\%), failing to effectively identify key instances. This behavior also reflects their poorer bag-level classification performance at lower witness rates. 
%Both results are visually summarized in Figure \ref{fig:res} (a)-(b).

\section{Discussion}

The experimental results highlight clear differences in how the evaluated methods respond to varying witness rates (WR). At moderate WRs (10\% and 5\%), most methods reach reasonable performance and with SLAM-AGS consistently achieving the highest bag-level F1-Score and competitive Top 400 detection accuracy, demonstrating that the combination of WeakSupCon and adaptive PCGrad effectively captures discriminative representations when sufficient positive supervision is available.

As the witness rate decreases to 1\% and 0.5\%, weakly supervised and self-supervised methods show a marked decline in performance. This drop can be attributed to the scarcity of informative positive instances within each bag. Self-supervised approaches like SimCLR also struggle to extract meaningful features. Hybrid methods such as WeakSupCon may suffer from conflicting gradients and unstable learning, resulting in poor generalization.

In contrast, SLAM-AGS maintains robust performance across all WRs. By projecting conflicting gradients and adaptively rescaling the combined update, it makes the supervised and self-supervised contrastive objectives reinforce rather than interfere, preserving discriminative features for rare positives even at sub-percent witness rates.

Interestingly, while the Top 400 detection accuracy of SLAM-AGS remains relatively stable across WRs, the corresponding bag-level F1 decreases as WR decreases. This suggests that although the model continues to identify relevant key instances, the used bag-level aggregator (PAMIL) struggles to correctly combine these signals under very sparse supervision.

%Overall, these findings emphasize that, under limited supervision, self-supervised or weakly supervised learning alone is insufficient to reliably capture rare positive instances. The integration of adaptive gradient conflict resolution, as implemented in SLAM-AGS, provides a robust and promising approach for maintaining meaningful representations and stable performance under scarce and noisy supervision.

\section{Conclusion}
We presented SLAM-AGS, a method that combines WeakSupCon with a novel adaptive gradient conflict resolution strategy. Our approach consistently outperforms self-supervised and 
weakly supervised methods across varying witness rates, maintaining both bag-level classification accuracy and instance-level detection under scarce supervision. These results demonstrate that, in a multi-task setting, mitigating gradient conflicts and rescaling updates are critical for robust representation learning when positive signals are sparse. Future work may focus on improving the aggregator to better leverage instance-level information and further enhance performance at extremely low witness rates.
\newpage
\vspace{1em}
\begin{center}
    {\bf\uppercase{Acknowledgements}}
\end{center}
This work is supported by the Swedish Cancer Society (Cancerfonden) through project 22 2353 Pj.
\vspace{1em}
\begin{center}
    {\bf\uppercase{Disclosure of Interests}}
\end{center}
The authors have no competing interests to declare that are relevant to the content of this article.

% suppress any automatic bibliography title
\renewcommand{\refname}{}%

% print a centered REFERENCES header with IEEE-like font/weight
\bibliographystyle{IEEEbib}
\bibliography{refs}

\end{document}